\documentclass[11pt]{article}

\usepackage[preprint]{acl}

\usepackage{times}
\usepackage{latexsym}

\usepackage[T1]{fontenc}
\usepackage[utf8]{inputenc}

\usepackage{microtype}

\usepackage{inconsolata}

\usepackage{graphicx}
\usepackage{tcolorbox}
\tcbuselibrary{skins}  % skins → borderline west; shadows → drop shadow
\usepackage{tabularx}
\usepackage{booktabs}     
\usepackage{longtable}
\usepackage{placeins}   % \FloatBarrier
\usepackage{stfloats}

\title{The Many Morals of AI: The cautionary bias of North American news narratives in global perspective}

\author{Andrew Piper$^{1}$ \\
  \texttt{andrew.piper@mcgill.ca}
  \And
  Edie Pearman$^{1}$ \\
  \texttt{edie.pearman@mail.mcgill.ca}
  \AND
  $^1$Department of Languages, Literatures, and Cultures \\
  McGill University, Montreal, QC H3A 2M7
}
  
\begin{document}
\maketitle
\nolinenumbers
\begin{abstract}
The stories told about artificial intelligence (AI) play a central role in shaping public understanding, policy debates, and expectations about technological change. News media are especially influential in this process. Using a cross-regional corpus of 12,924 English-language news articles about AI published since 2022, we examine how storytelling about AI varies across media systems in four regions: North America, China, India, and the Middle East. Leveraging large language models, we extract the implicit narrative lessons conveyed by each article—commonly known as \textit{story morals}—capturing value-laden interpretations that extend beyond thematic content or sentiment. We find that North American AI coverage is uniquely characterized by narratives of caution, while under-emphasizing narratives of innovation and collaboration that are more prevalent elsewhere. However, this pattern is largely explained by a broader regional negativity bias in North American news sentiment, rather than by heightened concern about AI itself. Our findings bring into view the culturally conditioned nature of AI news narratives. 
\end{abstract}

\section{Introduction}

% The prominent journalistic skepticism toward AI familiar to North American readers should not be mistaken for a global consensus around technological risk and reward. Rather, narratives around AI are shaped by regional media systems and local interpretive frameworks. As our research shows, no one AI narrative fits all.

The stories we tell about AI shape how individuals understand the technology's role in society. This in turn can shape how local publics interpret the balance of risk and reward that accompanies this emerging technology.

News media occupy a uniquely powerful position in the storytelling ecosystem of technological change \citep{walgrave2006contingency, boydstun2022making, wlezien2024media}. As recurring acts of sense-making, news stories help structure public attention by selecting which events count as ``news'' \citep{mccombs1972agenda} and framing them within moral and emotional registers of risk and reward \citep{groves2016ten, chong2007framing, iyengar1994anyone, cave2018portrayals}.

A substantial body of scholarship on AI's portrayal in the news has emerged in recent years. Early work prior to the emergence of chatbots highlighted that AI coverage focused far more on themes of productivity and efficiency than issues surrounding governance or ethical concerns \citep{chuan2019framing, sun2020newspaper, ouchchy2020ai}. When ethical questions have surfaced, they are often framed in practical or procedural terms rather than as moral debate \citep{ouchchy2020ai, bunz2022ai, nguyen2024news}. This oscillation has been characterized as a shift between utopian ``helping hand'' and dystopian ``Frankenstein'' frames \citep{cools2024exactly}. More recent work has highlighted a growing negativity surrounding AI coverage \citep{ryazanov2024chatgpt}, with North American news showing higher relative negativity compared to other regions \citep{ittefaq2025global}.

% Compact badges
\newtcbox{\pill}{on line,
  colback=gray!15, colframe=gray!50,
  boxrule=0.3pt, arc=2pt,
  left=2pt, right=2pt, top=0pt, bottom=0pt}

\newtcbox{\valencebadge}{on line,
  colback=red!10, colframe=red!60!black,
  boxrule=0.4pt, arc=2pt,
  left=3pt, right=3pt, top=0pt, bottom=0pt}

% === Compact framed summary table ===
\begin{figure}[t]
\centering
\begin{tcolorbox}[
  enhanced,
  width=\linewidth,
  colback=white,
  colframe=black,
  boxrule=0.5pt,
  borderline west={4pt}{0pt}{red!70!black},
  sharp corners,
  drop shadow,
  left=5pt, right=5pt,
  top=3pt, bottom=3pt,
  boxsep=2pt,
  fonttitle=\small,
  title={\textbf{Federal Court bans judges from using AI in decisions for time being}\\
         \footnotesize\textit{Source: The Globe and Mail}}
]
\small
\renewcommand{\arraystretch}{1.0}
\begin{tabularx}{\linewidth}{@{}>{\bfseries}l X@{}}
\toprule
Moral &
The cautious and transparent integration of artificial intelligence in judicial processes is essential to maintain public trust and ensure justice and fairness.\\
\addlinespace[2pt]
Keywords &
\pill{trust}\ \pill{justice}\ \pill{fairness} \\
\addlinespace[2pt]
Valence &
\valencebadge{\textbf{Negative}} \\
\bottomrule
\end{tabularx}
\end{tcolorbox}
\caption{Example of our story moral extraction process.}
\label{fig:moral_extraction}
\end{figure}

While prior work has made important contributions to our understanding of how AI is framed in the news, there are significant limitations to the data and methodologies used to date. Work in this area has been hampered by a lack of regional diversity, with only one large-scale study focusing on cross-cultural comparisons \citep{ittefaq2025global}. Several studies rely exclusively on qualitative analysis of small samples (N $<$ 1,000) \citep{bunz2022ai, ouchchy2020ai, chuan2019framing}, while computational analysis has typically relied on methods such as topic modeling and/or dictionaries \citep{sun2020newspaper, nguyen2023news, cools2024exactly, ittefaq2025global}, which have known limitations in terms of interpretive reliability \citep{mantyla2018measuring, hecking2019can, rijcken2024topic, pham2024topicgpt} and ignore underlying narrative structures of news articles. 

To address these limitations, we combine large-scale cross-regional analysis with recent advances in computational narrative understanding \citep{ranade2022computational}. Using a dataset of 12,924 news stories published since 2022 from four distinct regions—North America, the Middle East, China, and India—we apply methods for inferring the central values and lessons conveyed by a narrative, commonly described as \textit{story moral inference} (\autoref{fig:moral_extraction}). This approach has been validated across multiple narrative domains \citep{hobson2024story}, and has shown robustness across languages and cultural settings in capturing the higher-order lessons that readers infer from stories \citep{zhou2024large, wu2026lessons}. Although related to media framing, story-moral inference targets a different level of narrative meaning: it seeks to recover the implicit evaluative point of view of a story, in effect asking, ``Why was this story told?'' rather than identifying its topics, themes, or issue-specific emphases.

Using this approach, we find that North American news about AI exhibits a distinctive narrative profile, emphasizing narratives of caution more strongly than those of innovation or collaboration. Importantly, this pattern is strongly associated with a broader negativity bias in North American news relative to other regions, suggesting the important role that culturally-inflected media conventions play in journalistic reporting. Our findings underscore that the cautionary narratives prominent in North American news should not be mistaken for a global consensus about the technology's risks or social meaning.

\section{Data and Methods}

\subsection{Data Collection} 

For the purposes of our data collection, we utilized the Dow Jones Factiva platform limiting our queries between Jan. 1, 2022 (preceding chatGPT's launch) and Sept. 9, 2024. Article sources were limited to 12 major English-language news organizations across four regions: North America, China, India, and the Middle East (\autoref{tab:ai_article_counts}). 

As a first filtering step, we retrieved articles using two broad search terms (“Artificial Intelligence” OR “AI”) in order to capture the full spectrum of \textit{potential} AI discussion. This resulted in an initial collection of 48,201 articles. To distinguish articles that genuinely focused on AI from those that merely mentioned it or triggered false positives (e.g. `Air India'), we tested four automated filtering protocols using GPT-4o against a reference set of 100 manually labeled articles: i) identifying AI as the ``main character" or ``agent/object of opinions," (ii) detecting AI mentions in the first five sentences, (iii) counting total AI mentions, and (iv) extracting three central topics of the article and determining whether AI appeared in this set. Our best method (iv) showed a weighted F1 = 0.85, which when applied to the full dataset reduced the total number of articles to 8,924.

For comparison purposes, we then randomly selected 1,000 articles from each region (N=4,000) drawn from the rejected set of initially downloaded articles. This selection method helps guarantee the greatest alignment between our two primary article types (about AI, not about AI) while controlling for sections, length, industry focus and other confounding factors. 

% We first created a test set of 100 randomly selected articles which were manually labeled by one of the authors as being ``about AI" or not. We then tested six automated classification methods using GPT-4o: i) identifying AI as the ``main character" or ``agent/object of opinions," (ii) detecting AI mentions in the first five sentences, (iii) counting total AI mentions, and (iv) extracting three central topics of the article and determining whether AI appeared in this set. The three central topics method (iv) performed best (weighted F1 = 0.85, with 78\% recall and 81\% precision). This filtering step yielded a final dataset of 8,924 AI-related articles, notably smaller than the initial keyword filtering approach. 

\subsection{Inferring Story Morals using LLMs}
\label{sec:Moral_Method}
For all articles, we prompted GPT-4o and Gemini-2.0-Flash to identify the following: (i) story moral statement in the form of a single sentence, (ii) 2-3 keywords distilled from the moral sentence, (iii) sentiment valence, and (iv) candidate industries mentioned in the article. (The full prompting notebooks are included in \autoref{app:prompt}.) Given the sensitivity of LLM outputs to model architecture and prompt variation shown in prior work \citep{sclar2024quantifying, baumann2025large}, we generate eight prompt variations manipulating phrasing and position of article text. For each article, we then randomly select one prompt variation and submit it to both models. We then aggregate both models' outputs as our final dataset.\footnote{The extrapolated rank-biased overlap (RBO) between the GPT-4o and Gemini rankings of keywords is 0.32 for the top 10 keywords and 0.40 for the top 100 ($p = 0.9$) indicating low to moderate agreement.}

To validate moral keyword quality, we randomly sampled 100 AI articles and divided them into two groups of 50. For each group, two independent human annotators rated keyword relevance (0 = not relevant, 1 = moderately relevant, 2 = highly relevant) without knowledge of which model generated the keywords. We averaged the two scores for each article. Overall, a strong majority of keywords from both models (GPT-4o: 96\%, Gemini: 82\%) achieved at least moderate relevance, justifying their inclusion in subsequent analysis.

For estimation of article-level valence, we follow recent work and apply our randomized prompting framework to the generation of an article-level three-class sentiment framework: negative, neutral, and positive \citep{zhang2024sentiment, debess2024good}.

\section{Results}\label{sec2}

\subsection{Distribution of moral keywords by region}

\begin{figure}[h]
\centering
\includegraphics[width=.8\linewidth]{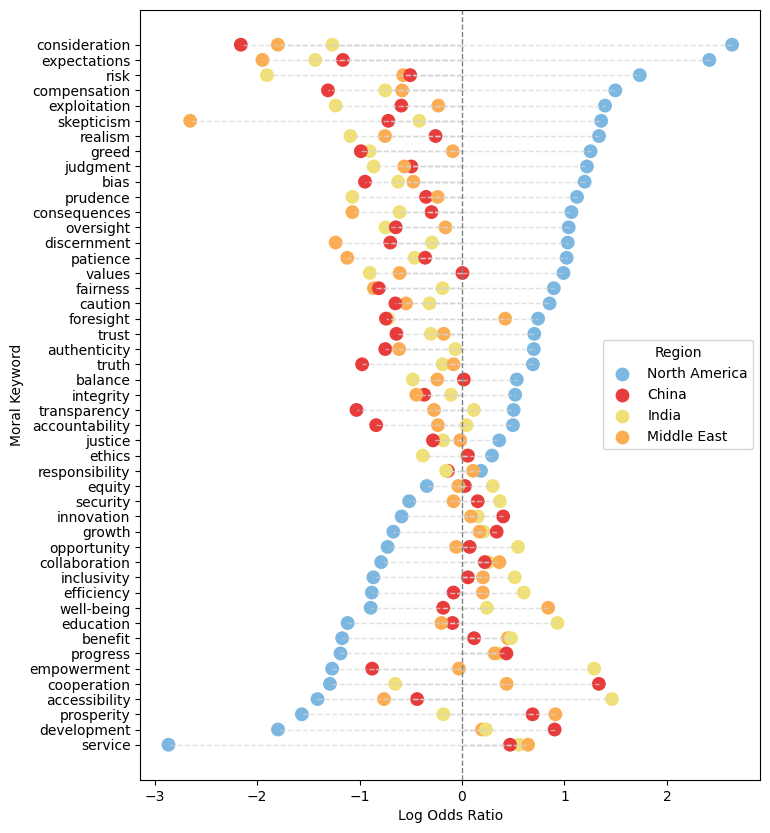}
\caption{Moral keywords most strongly / weakly associated with North American news stories.}
\label{fig:NA_keywords}
\end{figure}

We begin by identifying the moral vocabularies that distinguish North American AI coverage from other regions, using log-odds ratios to compare moral keywords in North American stories with those in the rest of the corpus. We find that North American coverage disproportionately emphasizes \textit{skepticism, risk, exploitation, greed,} and \textit{bias}, while placing less emphasis on innovation-related values such as \textit{development, prosperity, progress,} and \textit{competitiveness}, as well as social values such as \textit{collaboration, community,} and \textit{cooperation} (\autoref{fig:NA_keywords}). North American AI storytelling thus foregrounds risk while giving less attention to collective benefit and shared development.

\begin{figure}[h]
    \centering
    \includegraphics[width=0.7\linewidth]{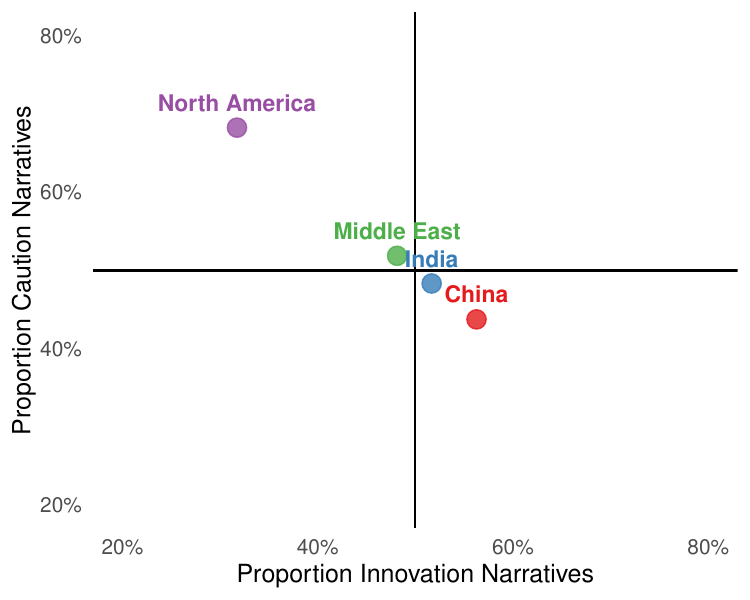}
    \caption{Regional variation in the balance between caution and innovation narratives, which represent 50\% of the total documents in our corpus.}
    \label{fig:narrative_regions}
\end{figure}

To clarify these divisions more sharply, we organize moral keywords into two overarching narrative types: \textit{caution narratives}, which center on risk, responsibility, ethics, regulation, and harm, and \textit{innovation narratives}, which focus on progress, opportunity, efficiency, and growth (\autoref{fig:narrative_regions}, see \autoref{app:caution_coding} for process). 

% One important detail worth noting: the strong North American orientation towards caution does not imply that other regions are exclusively focused on narratives of innovation. Rather, we see more balanced levels between these competing narrative frames across the three other regions, suggesting a general awareness of risk that nevertheless does not rise to the dominant framework. 

\subsection{The association of sentiment and cautionary AI framing}
\label{sec:sentiment_caution}

\begin{figure}[h!]
  \centering
  \includegraphics[width=.9\linewidth]{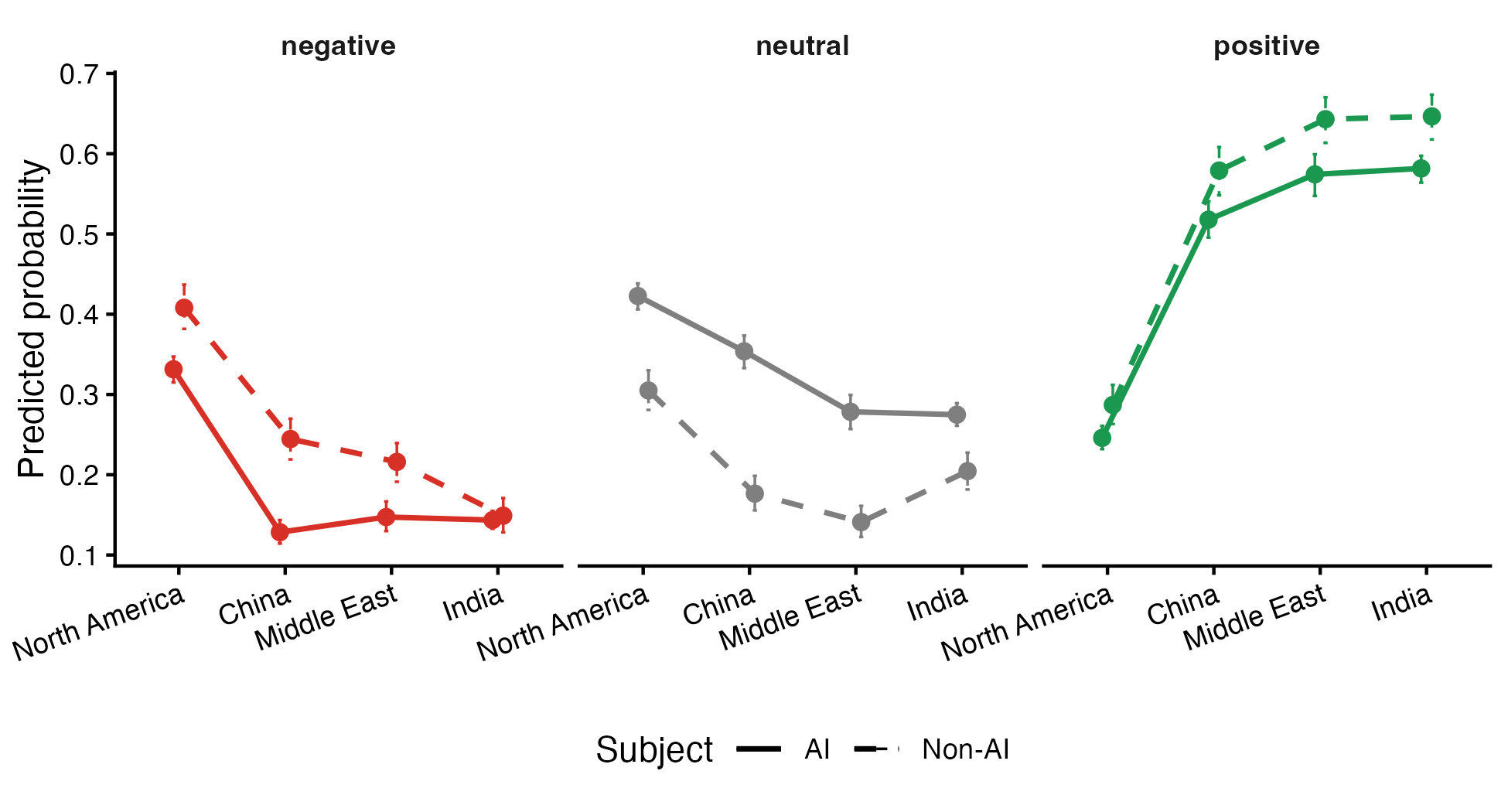}
  \caption{Predicted probabilities of negative, neutral, and positive sentiment by article content from a multinomial logistic regression including region, article subject, their interaction, annotation model, and categorical prompt identity.}
  \label{fig:valence_mediation}
\end{figure}

One possible explanation for the distinctively cautionary character of North American AI coverage is that it reflects a unique concern about AI. Another is that it reflects a more general tendency within North American news to narrate events through negative frames. To distinguish between these possibilities, we fitted a multinomial logistic regression of sentiment (Negative, Neutral, Positive) within articles, including  region, subject group (AI / not-AI), their interaction, and categorical controls for annotation model and prompt (\autoref{tab:sentiment_multinom}). We found that North American articles were significantly more likely to be negative in both groups: predicted probabilities were 33.1\% for AI-focused articles and 40.8\% for comparison articles, versus 12.8--14.7\% and 14.9--24.5\%, respectively, in other regions (\autoref{fig:valence_mediation}).

We next asked whether North American AI coverage remains more likely to convey cautionary morals after accounting for its more negative sentiment. For each of the 17,848 model--prompt annotations of only the AI articles, we combined all extracted moral keywords and coded whether at least one belonged to our cautionary framework. We fitted binary logistic regressions predicting this outcome from region, annotation model, and prompt, before and after adding categorical sentiment, again clustering uncertainty by article (\autoref{tab:caution_logit}). We found that the North America--other regions gap declined from 15.1 percentage points before sentiment adjustment to 5.0 points afterward. The adjusted gap remained positive when each newspaper was omitted in turn (3.9--6.3 points) and in separate analyses of both annotation models. North America’s greater prevalence of cautionary AI morals is thus substantially reduced, though not entirely eliminated, when regional sentiment differences are taken into account and different sources and models are measured. This suggests that this cautionary framing is associated with a broader pattern of negative sentiment while also exhibiting a regional difference that sentiment alone does not fully capture.

\subsection{Negativity Around AI Is Unevenly Distributed Across Industries} 
\begin{figure}[h!]
\centering
\includegraphics[width=.8\linewidth]{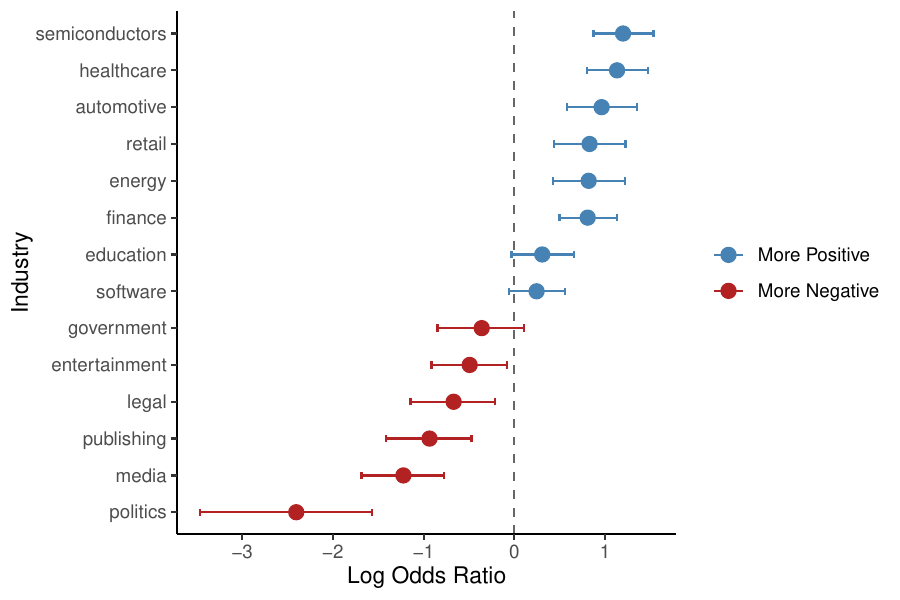}
\caption{Log odds ratios (OR) of positive valence in North American news by industry for the top 15 industries.}
\label{fig:valence_industry}
\end{figure}

We examined whether the negativity observed in North American media coverage was concentrated in particular industries by calculating odds ratios (ORs) of positive versus negative valence for the top 15 industries in this subset of the data. Log-transformed odds ratios (log ORs) and 95\% confidence intervals (CIs) are shown in \autoref{fig:valence_industry}. Negatively valenced industries cluster around media and entertainment (along with law), with politics exhibiting the highest level of concern. Other industries such as healthcare, semiconductors, retail and finance were generally more positively portrayed. 

% These findings demonstrate that the negativity in North American media around AI is not uniform but concentrated within a subset of particular industries which are more likely to deal in symbolic capital as well as regulatory power.

\section{Conclusion} 

Our findings suggest that narratives surrounding AI in the news reflect not only local evaluations of the technology but also regional conventions of broader journalistic storytelling. Although North American coverage of AI disproportionately emphasizes cautionary values, this distinctiveness is substantially reduced when sentiment differences are taken into account. The greater negativity of North American coverage in general extends beyond articles centrally focused on AI, suggesting that cautionary storytelling belongs to a broader regional pattern. These findings suggest that public-facing narratives about AI are embedded within deeper media cultures of evaluation and affect, with potentially important consequences for how technological change is collectively understood. 

Our results help connect two strands of prior research on AI coverage that have often been treated separately: work documenting the growing negativity of AI-related news, particularly in North America \citep{ryazanov2024chatgpt}, and work emphasizing regional variation in media cultures \citep{ittefaq2025global}. Our analysis shows that the moral coloration of AI storytelling is closely associated with regional patterns of sentiment. In contrast to journalistic cultures elsewhere in the world that balance narratives of caution \textit{and} innovation and collaboration, the entrenched negativity of North American news media strongly skews AI storytelling toward cautionary frames. That being said, our findings also show that narrative values, while strongly associated with sentiment, are not completely reducible to sentiment. 

% The concentration of negative AI sentiment within particular industries further refines this picture. North American negativity is not uniformly distributed across AI applications but is instead clustered in domains tied to symbolic authority, creativity, and governance (media, law, and politics). By contrast, sectors such as healthcare, finance, and manufacturing are framed more positively, often in terms of efficiency and progress. This suggests that moral narratives about AI are more responsive to perceived threats to cultural value and civic institutions rather than  technical capacity.

% Methodologically, our brings to the fore the value of narrative-centered approaches to large-scale media analysis. Traditional topic models and sentiment dictionaries capture what news stories are about and how they feel, but they often miss the underlying normative questions that motivate storytelling: why a story is told and what lessons it conveys. By extracting high-level narratological concepts such as ``story morals,'' we are able to surface another interpretive layer around which media framing shapes public meaning.

Overall, our work illustrates the importance of cross-regional comparison in understanding how technological change is narrated. For journalists, the findings invite reflection on whether the moral lessons conveyed by AI follow from the evidence of a particular story or draw on familiar sentiment conventions. For policymakers, they underscore the need to distinguish the narratives prominent in one's own media environments from a global assessment of AI’s risks and benefits. Recognizing these regional patterns can bring a wider range of perspectives into public debate.

\section*{Limitations}

Several limitations should be noted in conclusion. Our analysis relies on large language models to infer story morals, which introduces dependence on model interpretive priors. While prior work has shown that frontier models like GPT and Gemini used here align with human judgments and are often preferred in blind surveys with respect to story moral quality \citep{hobson2024story}, our findings are nevertheless limited by the semantic frameworks of models. Even though we control for model and prompt effects and focus on robust aggregate patterns, future work should further validate moral inference and the potential interpretive biases introduced by LLMs. In addition, our regional categories necessarily compress internal diversity within media systems, and our findings should not be read as exhaustive characterizations of national or cultural attitudes toward AI. Our sample was designed with cross-cultural comparison first and foremost in mind. Additionally, mediation analyses in observational settings do not establish causal pathways: they provide evidence consistent with sentiment operating as an important explanatory factor linking region and moral framing. Future work can investigate this relationship between systemic valence and narrative generation at the publication or article level more deeply. Finally, while we assume the distribution of news narratives is an on-going dynamic process that may change after our data cut-off, our findings also suggest a strong anchoring via regional sentiment distributions which may make the story morals of AI more consistent than expected.

\section*{Data Availability}
A dataset of 12,924 news articles from 12 different sources is made available along with annotations for story morals, keywords, valence, and industry. \url{https://anonymous.4open.science/r/AI-in-the-news-3712/}

\section*{AI Disclosure Statement}
OpenAI Codex assisted with idea development, outlining, language editing, code development, and data analysis. The authors reviewed and revised its contributions and take full responsibility for the final manuscript, analyses, and conclusions.

%\section*{Acknowledgments}

% Author contributions: A.P. and E.P. designed research, E.P. performed research, A.P. and E.P. analyzed data, and A.P. and E.P. wrote the paper.}

% Bibliography entries for the entire Anthology, followed by custom entries
%\bibliography{custom,anthology-overleaf-1,anthology-overleaf-2}

% Custom bibliography entries only
\bibliography{custom}

@inproceedings{zhang2024sentiment,
  title={Sentiment analysis in the era of large language models: A reality check},
  author={Zhang, Wenxuan and Deng, Yue and Liu, Bing and Pan, Sinno and Bing, Lidong},
  booktitle={Findings of the Association for Computational Linguistics: NAACL 2024},
  pages={3881--3906},
  year={2024}
}

@inproceedings{debess2024good,
  title={Good or bad news? Exploring GPT-4 for sentiment analysis for faroese on a public news corpora},
  author={Debess, Iben Nyholm and Simonsen, Annika and Einarsson, Hafsteinn},
  booktitle={Proceedings of the 2024 Joint International Conference on Computational Linguistics, Language Resources and Evaluation (LREC-COLING 2024)},
  pages={7814--7824},
  year={2024}
}

@inproceedings{sclar2024quantifying,
  title={Quantifying Language Models' Sensitivity to Spurious Features in Prompt Design or: How I learned to start worrying about prompt formatting},
  author={Sclar, Melanie and Choi, Yejin and Tsvetkov, Yulia and Suhr, Alane},
  booktitle={International Conference on Learning Representations},
  volume={2024},
  pages={25055--25083},
  year={2024}
}

@article{baumann2025large,
  title={Large language model hacking: Quantifying the hidden risks of using llms for text annotation},
  author={Baumann, Joachim and R{\"o}ttger, Paul and Urman, Aleksandra and Wendsj{\"o}, Albert and Plaza-del-Arco, Flor Miriam and Gruber, Johannes B and Hovy, Dirk},
  journal={arXiv preprint arXiv:2509.08825},
  year={2025}
}

@inproceedings{zhou2024large,
  title={Large scale narrative messaging around climate change: A cross-cultural comparison},
  author={Zhou, Haiqi and Hobson, David G and Ruths, Derek and Piper, Andrew},
  booktitle={Proceedings of the 1st Workshop on Natural Language Processing Meets Climate Change (ClimateNLP 2024)},
  pages={143--155},
  year={2024}
}

@article{wu2026lessons,
  title={Lessons Without Borders? Evaluating Cultural Alignment of LLMs Using Multilingual Story Moral Generation},
  author={Wu, Sophie and Piper, Andrew},
  journal={arXiv preprint arXiv:2604.08797},
  year={2026}
}

@inproceedings{hobson2024story,
  title={Story Morals: Surfacing value-driven narrative schemas using large language models},
  author={Hobson, David and Zhou, Haiqi and Ruths, Derek and Piper, Andrew},
  booktitle={Proceedings of the 2024 Conference on Empirical Methods in Natural Language Processing},
  pages={12998--13032},
  year={2024}
}

@article{ranade2022computational,
  title={Computational understanding of narratives: A survey},
  author={Ranade, Priyanka and Dey, Sanorita and Joshi, Anupam and Finin, Tim},
  journal={IEEE Access},
  volume={10},
  pages={101575--101594},
  year={2022},
  publisher={IEEE}
}

@inproceedings{mantyla2018measuring,
  title={Measuring LDA topic stability from clusters of replicated runs},
  author={Mantyla, Mika V and Claes, Maelick and Farooq, Umar},
  booktitle={Proceedings of the 12th ACM/IEEE international symposium on empirical software engineering and measurement},
  pages={1--4},
  year={2018}
}

@inproceedings{pham2024topicgpt,
  title={TopicGPT: A Prompt-based Topic Modeling Framework},
  author={Pham, Chau and Hoyle, Alexander and Sun, Simeng and Resnik, Philip and Iyyer, Mohit},
  booktitle={Proceedings of the 2024 Conference of the North American Chapter of the Association for Computational Linguistics: Human Language Technologies (Volume 1: Long Papers)},
  pages={2956--2984},
  year={2024}
}

@article{rijcken2024topic,
  title={Topic specificity: A descriptive metric for algorithm selection and finding the right number of topics},
  author={Rijcken, Emil and Zervanou, Kalliopi and Mosteiro, Pablo and Scheepers, Floortje and Spruit, Marco and Kaymak, Uzay},
  journal={Natural Language Processing Journal},
  pages={100082},
  year={2024},
  publisher={Elsevier}
}

@article{hecking2019can,
  title={Can topic models be used in research evaluations? Reproducibility, validity, and reliability when compared with semantic maps},
  author={Hecking, Tobias and Leydesdorff, Loet},
  journal={Research Evaluation},
  volume={28},
  number={3},
  pages={263--272},
  year={2019},
  publisher={Oxford University Press}
}

@article{ittefaq2025global,
  title={Global news media coverage of artificial intelligence (AI): A comparative analysis of frames, sentiments, and trends across 12 countries},
  author={Ittefaq, Muhammad and Zain, Ali and Arif, Rauf and Ala-Uddin, Mohammad and Ahmad, Taufiq and Iqbal, Azhar},
  journal={Telematics and Informatics},
  volume={96},
  pages={102223},
  year={2025},
  publisher={Elsevier}
}

@article{ryazanov2024chatgpt,
  title={How ChatGPT changed the media’s narratives on AI: a semi-automated narrative analysis through frame semantics},
  author={Ryazanov, Igor and {\"O}hman, Carl and Bj{\"o}rklund, Johanna},
  journal={Minds and Machines},
  volume={35},
  number={1},
  pages={2},
  year={2024},
  publisher={Springer}
}

@article{cave2018portrayals,
  title={Portrayals and perceptions of AI and why they matter},
  author={Cave, Stephen and Craig, Claire and Dihal, Kanta and Dillon, Sarah and Montgomery, Jessica and Singler, Beth and Taylor, Lindsay},
  year={2018},
  publisher={The Royal Society}
}

@article{bunz2022ai,
  title={The AI doctor will see you now: assessing the framing of AI in news coverage},
  author={Bunz, Mercedes and Braghieri, Marco},
  journal={AI \& society},
  volume={37},
  number={1},
  pages={9--22},
  year={2022},
  publisher={Springer}
}

@article{ouchchy2020ai,
  title={AI in the headlines: the portrayal of the ethical issues of artificial intelligence in the media},
  author={Ouchchy, Leila and Coin, Allen and Dubljevi{\'c}, Veljko},
  journal={AI \& society},
  volume={35},
  number={4},
  pages={927--936},
  year={2020},
  publisher={Springer}
}

@article{nguyen2023news,
  title={How news media frame data risks in their coverage of big data and AI},
  author={Nguyen, Dennis},
  journal={Internet Policy Review},
  volume={12},
  number={2},
  pages={1--30},
  year={2023},
  publisher={Berlin: Alexander von Humboldt Institute for Internet and Society}
}

@article{cools2024exactly,
  title={Where exactly between utopia and dystopia? A framing analysis of AI and automation in US newspapers},
  author={Cools, Hannes and Van Gorp, Baldwin and Opgenhaffen, Michael},
  journal={Journalism},
  volume={25},
  number={1},
  pages={3--21},
  year={2024},
  publisher={SAGE Publications Sage UK: London, England}
}

@article{sun2020newspaper,
  title={Newspaper coverage of artificial intelligence: A perspective of emerging technologies},
  author={Sun, Shaojing and Zhai, Yujia and Shen, Bin and Chen, Yibei},
  journal={Telematics and Informatics},
  volume={53},
  pages={101433},
  year={2020},
  publisher={Elsevier}
}

@article{nguyen2024news,
  title={The news framing of artificial intelligence: a critical exploration of how media discourses make sense of automation},
  author={Nguyen, Dennis and Hekman, Erik},
  journal={AI \& society},
  volume={39},
  number={2},
  pages={437--451},
  year={2024},
  publisher={Springer}
}

@inproceedings{chuan2019framing,
  title={Framing artificial intelligence in American newspapers},
  author={Chuan, Ching-Hua and Tsai, Wan-Hsiu Sunny and Cho, Su Yeon},
  booktitle={Proceedings of the 2019 AAAI/ACM Conference on AI, Ethics, and Society},
  pages={339--344},
  year={2019}
}

@article{mccombs1972agenda,
  title={The agenda-setting function of mass media},
  author={McCombs, Maxwell E and Shaw, Donald L},
  journal={Public opinion quarterly},
  volume={36},
  number={2},
  pages={176--187},
  year={1972},
  publisher={Oxford University Press}
}

@book{iyengar1994anyone,
  title={Is anyone responsible?: How television frames political issues},
  author={Iyengar, Shanto},
  year={1994},
  publisher={University of Chicago Press}
}

@article{chong2007framing,
  title={Framing theory},
  author={Chong, Dennis and Druckman, James N},
  journal={Annu. Rev. Polit. Sci.},
  volume={10},
  number={1},
  pages={103--126},
  year={2007},
  publisher={Annual Reviews}
}

@article{walgrave2006contingency,
  title={The contingency of the mass media's political agenda setting power: Toward a preliminary theory},
  author={Walgrave, Stefaan and Van Aelst, Peter},
  journal={Journal of communication},
  volume={56},
  number={1},
  pages={88--109},
  year={2006},
  publisher={Oxford University Press}
}

@article{wlezien2024media,
  title={Media reflect! Policy, the public, and the news},
  author={Wlezien, Christopher and Soroka, Stuart},
  journal={American Political Science Review},
  volume={118},
  number={3},
  pages={1563--1569},
  year={2024},
  publisher={Cambridge University Press}
}

@book{boydstun2022making,
  title={Making the news: Politics, the media \& agenda setting},
  author={Boydstun, Amber E},
  year={2022},
  publisher={University of Chicago Press}
}

@article{groves2016ten,
  title={Ten years of science news: A longitudinal analysis of scientific culture in the Spanish digital press},
  author={Groves, Tamar and Figuerola, Carlos G and Quintanilla, Miguel {\'A}},
  journal={Public Understanding of Science},
  volume={25},
  number={6},
  pages={691--705},
  year={2016},
  publisher={SAGE Publications Sage UK: London, England}
}

\appendix

% Sections: A1, A2, A3 ...
\renewcommand{\thesection}{A\arabic{section}}

% Subsections: A1.1, A1.2, A2.1 ...
\renewcommand{\thesubsection}{\thesection.\arabic{subsection}}

% Make autoref say "Appendix A1"
\renewcommand{\sectionautorefname}{Appendix}
\renewcommand{\subsectionautorefname}{Appendix}

% Appendix tables should be numbered A1, A2, A3...
\setcounter{table}{0}
\renewcommand{\thetable}{A\arabic{table}}

\section*{Appendix}

\section{Article Counts by Publication and Region}

\begin{table}[h]
\centering
\small
\begin{tabular}{@{}llrr@{}}
\toprule
\textbf{Source Name} & \textbf{Region} & \textbf{AI} & \textbf{Non-AI} \\
\midrule
South China Morning Post   & China      & 986   & 423 \\
China Daily                & China      & 700   & 577 \\
The Times of India         & India      & 1,766 & 493 \\
Hindustan Times            & India      & 902   & 315 \\
The Hindu                  & India      & 499   & 192 \\
The National               & M. East    & 638   & 372 \\
Gulf News (Dubai)          & M. East    & 376   & 429 \\
Al Jazeera English         & M. East    & 225   & 199 \\
The Wall Street Journal    & N. America & 1,112 & 403 \\
The New York Times         & N. America & 721   & 232 \\
The Washington Post        & N. America & 570   & 194 \\
The Globe and Mail         & N. America & 429   & 171 \\
\bottomrule
\end{tabular}
\caption{Article Counts by News Source, Region, and Category}
\label{tab:ai_article_counts}
\end{table}

\section{Coefficient Tables}
\label{app:full-model-tables}

Coefficients for the models reported in Section~\ref{sec:sentiment_caution}. \autoref{tab:sentiment_multinom} reports the multinomial logistic regression of sentiment, and \autoref{tab:caution_logit} the logistic regressions of cautionary framing before and after sentiment adjustment.

\begin{table*}[t]
\centering
\small
\begin{tabular}{@{}lrrrrrr@{}}
\toprule
 & \multicolumn{3}{c}{Neutral vs. Negative} & \multicolumn{3}{c}{Positive vs. Negative} \\
\cmidrule(lr){2-4} \cmidrule(l){5-7}
Term & Estimate & SE & $p$ & Estimate & SE & $p$ \\
\midrule
Intercept & $-$0.045 & 0.070 & .519 & $-$0.709 & 0.075 & $<$.001 \\
Region: China & 0.775 & 0.082 & $<$.001 & 1.702 & 0.086 & $<$.001 \\
Region: Middle East & 0.398 & 0.093 & $<$.001 & 1.669 & 0.093 & $<$.001 \\
Region: India & 0.411 & 0.067 & $<$.001 & 1.708 & 0.069 & $<$.001 \\
Subject: Non-AI & $-$0.536 & 0.077 & $<$.001 & $-$0.055 & 0.088 & .534 \\
China $\times$ Non-AI & $-$0.806 & 0.136 & $<$.001 & $-$0.479 & 0.136 & $<$.001 \\
Middle East $\times$ Non-AI & $-$0.529 & 0.149 & $<$.001 & $-$0.216 & 0.141 & .128 \\
India $\times$ Non-AI & 0.203 & 0.136 & .135 & 0.124 & 0.134 & .357 \\
Model: GPT-4o & 0.141 & 0.021 & $<$.001 & 0.286 & 0.014 & $<$.001 \\
Prompt 2 & 0.108 & 0.088 & .220 & 0.130 & 0.088 & .142 \\
Prompt 3 & 0.381 & 0.089 & $<$.001 & 0.360 & 0.090 & $<$.001 \\
Prompt 4 & 0.197 & 0.092 & .032 & 0.464 & 0.090 & $<$.001 \\
Prompt 5 & 0.207 & 0.088 & .018 & 0.165 & 0.088 & .062 \\
Prompt 6 & 0.136 & 0.090 & .129 & 0.326 & 0.089 & $<$.001 \\
Prompt 7 & 0.498 & 0.089 & $<$.001 & 0.382 & 0.090 & $<$.001 \\
Prompt 8 & 0.249 & 0.090 & .006 & 0.318 & 0.090 & $<$.001 \\
\bottomrule
\end{tabular}
\caption{Multinomial logistic regression of article sentiment (25,848 annotations of 12,924 articles). Negative sentiment is the outcome reference. Coefficients are log odds. Standard errors are clustered by article (HC1). Reference categories: North America, AI-focused articles, Gemini-2.0-Flash, and prompt 1.}
\label{tab:sentiment_multinom}
\end{table*}

\begin{table*}[t]
\centering
\small
\begin{tabular}{@{}lrrrrrr@{}}
\toprule
 & \multicolumn{3}{c}{Unadjusted} & \multicolumn{3}{c}{Sentiment-adjusted} \\
\cmidrule(lr){2-4} \cmidrule(l){5-7}
Term & Estimate & SE & $p$ & Estimate & SE & $p$ \\
\midrule
Intercept & 0.632 & 0.062 & $<$.001 & 1.207 & 0.071 & $<$.001 \\
Region: China & $-$0.780 & 0.054 & $<$.001 & $-$0.430 & 0.057 & $<$.001 \\
Region: Middle East & $-$0.454 & 0.061 & $<$.001 & $-$0.019 & 0.063 & .769 \\
Region: India & $-$0.663 & 0.046 & $<$.001 & $-$0.240 & 0.049 & $<$.001 \\
Model: GPT-4o & $-$0.727 & 0.022 & $<$.001 & $-$0.704 & 0.024 & $<$.001 \\
Prompt 2 & 0.000 & 0.077 & .998 & 0.068 & 0.078 & .383 \\
Prompt 3 & 0.158 & 0.077 & .040 & 0.263 & 0.078 & $<$.001 \\
Prompt 4 & 0.034 & 0.076 & .653 & 0.200 & 0.078 & .011 \\
Prompt 5 & 0.119 & 0.077 & .120 & 0.202 & 0.077 & .009 \\
Prompt 6 & $-$0.048 & 0.078 & .534 & 0.046 & 0.080 & .567 \\
Prompt 7 & 0.195 & 0.075 & .009 & 0.313 & 0.076 & $<$.001 \\
Prompt 8 & 0.074 & 0.076 & .326 & 0.189 & 0.077 & .014 \\
Sentiment: Neutral &  &  &  & $-$0.605 & 0.053 & $<$.001 \\
Sentiment: Positive &  &  &  & $-$1.648 & 0.053 & $<$.001 \\
\bottomrule
\end{tabular}
\caption{Logistic regressions of cautionary moral framing in AI-focused articles (17,848 annotations of 8,924 articles), without and with sentiment. The outcome is whether any extracted moral keyword is coded as caution. Coefficients are log odds. Standard errors are clustered by article (HC1). Reference categories: North America, Gemini-2.0-Flash, prompt 1, and negative sentiment.}
\label{tab:caution_logit}
\end{table*}

\FloatBarrier

\section{Caution / Innovation Framing Exercise}
\label{app:caution_coding}

% Keep this short; we can expand once you paste your mapping rules.
We coded whether each story’s moral framing emphasized \textit{caution} (risk, restraint, harms, governance concerns) and/or \textit{innovation} (progress, opportunity, productivity, benefits). A story could contain neither, one, or both framings.

\begin{table}[h]
\centering
\small
\begin{tabular}{p{0.25\columnwidth} p{0.25\columnwidth}}
\toprule
Caution & Innovation \\
\midrule
accountability & adaptation \\
caution & adaptability \\
ethics & creativity \\
fairness & efficiency \\
justice & empowerment \\
privacy & growth \\
prudence & innovation \\
responsibility & opportunity \\
risk & progress \\
safety & resilience \\
security & \\
transparency & \\
trust & \\
\bottomrule
\end{tabular}
\caption{Moral keywords mapped to caution and innovation framing categories.
\label{tab:keyword_mapping}}
\end{table}

\FloatBarrier

\section{Prompt Notebooks}\label{app:prompt}

All models were prompted using a temperature of 1.0 with a max token limit of 512.

%\vspace{1emedie.}

% EXPLICIT PROMPTS

% Topic Identification for filtering the articles for AI Relevance

\noindent \textbf{Topic Prompt.} This was used to filter non-AI related articles.
\begin{tcolorbox}[colback=gray!10, colframe=gray!10, boxrule=0pt]
\small
What are the central topics of the attached text? Provide your answer as a list of keywords separated by commas. Start from the most general topic and get increasingly specific. Give three total topics. Make sure your answer is just a simple list of keywords that are succinct. Do not combine themes into a single answer by using the word and. Choose one per item. You 
may use a multiword phrase however. Provide your answer as a list of three keywords separated by commas. Here is the passage:
\end{tcolorbox}

\noindent \textbf{Prompt Permutation Template.} To reduce sensitivity to prompt wording for moral statements, keywords, industries candidates, and sentiment valence, each prompt was drawn from a set of eight variations. These variations are created by using two phrasings of the target question, altering the position of the article text relative to the target question, and the inclusion or exclusion of a role statement: 
\begin{enumerate}
\item {[} PHRASING 1{]} Here is the text: {[}ARTICLE TEXT{]}
\item {[} PHRASING 2{]} Here is the text: {[}ARTICLE TEXT{]}
\item Here is the text: {[}ARTICLE TEXT{]} {[} PHRASING 1{]}
\item Here is the text: {[}ARTICLE TEXT{]} {[} PHRASING 2{]}
\item {[}ROLE STATEMENT{]} {[} PHRASING 1{]} Here is the text: {[}ARTICLE TEXT{]}
\item {[}ROLE STATEMENT{]} {[} PHRASING 2{]} Here is the text: {[}ARTICLE TEXT{]}
\item {[}ROLE STATEMENT{]} Here is the text: {[}ARTICLE TEXT{]} {[} PHRASING 1{]}
\item {[}ROLE STATEMENT{]} Here is the text: {[}ARTICLE TEXT{]} {[} PHRASING 2{]}
\end{enumerate}
When included, the role statement was either the general role statement or the moral keyword specific role statement. These role statements, as well as the various prompt phrasings, are provided below.

\vspace{1em}

\noindent \textbf{General Role Statement:}
\begin{tcolorbox}[colback=gray!10, colframe=gray!10, boxrule=0pt]
\small
Today, you are an expert story interpreter. I will give you a news article and ask you a question about it.
\end{tcolorbox}

%%% VALENCE
% Valence, 8 variations, 2 phrasings x before or after the article text x with or without role statement
\noindent \textbf{Valence Prompt, Phrasing 1:}
\begin{tcolorbox}[colback=gray!10, colframe=gray!10, boxrule=0pt]
\small
Determine the emotional tone or evaluative stance of AI in the story. State your answer as a single word only. \\
Your answer must be one of the following: 'positive', 'negative', or 'neutral'. \\
Please respond only with a valid JSON object in the following format: \\
\{\{ \\
"AI\_VALENCE": "\textless valence\textgreater" \\
\}\} \\
\end{tcolorbox}

\noindent \textbf{Valence Prompt, Phrasing 2:}
\begin{tcolorbox}[colback=gray!10, colframe=gray!10, boxrule=0pt]
\small
Is AI positively, negatively, or neutrally represented in this story? State your answer as a single word only. \\
Your answer must be one of the following: 'positive', 'negative', or 'neutral'. \\
Please respond only with a valid JSON object in the following format: \\
\{\{ \\
"AI\_VALENCE": "\textless valence\textgreater" \\
\}\} \\
\end{tcolorbox}

%%% INDUSTRY KEYWORDS
% Industry Keyword Prompt, again 8 variations, 2 phrasings x before or after the article text x with or without role statment

\noindent \textbf{Industry Prompt, Phrasing 1:}
\begin{tcolorbox}[colback=gray!10, colframe=gray!10, boxrule=0pt]
\small
What is the principal industry or industries in this story? State your answer as a list of single words. 
Please respond only with a valid JSON object in the following format: \\
\{\{ \\
"INDUSTRY\_1": "\textless industry\_name\textgreater", \\
"INDUSTRY\_2": "\textless industry\_name\textgreater", \\
... \\
\}\} \\
\end{tcolorbox}

\noindent \textbf{Industry Prompt, Phrasing 2:}
\begin{tcolorbox}[colback=gray!10, colframe=gray!10, boxrule=0pt]
\small
List the principal industry or industries important to this story. State your answer as a list of single words. 
Please respond only with a valid JSON object in the following format: \\
\{\{ \\
"INDUSTRY\_1": "\textless industry\_name\textgreater", \\
"INDUSTRY\_2": "\textless industry\_name\textgreater", \\
... \\
\}\} \\
\end{tcolorbox}

%%% MORAL STATEMENT
% Moral Statement Prompt, again 8 variations, 2 phrasings x before or after the article text x with or without role statement

\noindent \textbf{Moral Statement Prompt, Phrasing 1:}
\begin{tcolorbox}[colback=gray!10, colframe=gray!10, boxrule=0pt]
\small
What is the moral of this story? State your answer as a single sentence.\\
Please respond only with a valid JSON object in the following format:\\
\}\} \\
"MORAL\_SENTENCE": "\textless moral\textgreater"\\
\}\} \\
\end{tcolorbox}

\noindent \textbf{Moral Statement Prompt, Phrasing 2:}
\begin{tcolorbox}[colback=gray!10, colframe=gray!10, boxrule=0pt]
\small
How might one interpret the moral of this story? State your answer as a single sentence.\\
Please respond only with a valid JSON object in the following format:\\
\}\} \\
"MORAL\_SENTENCE": "\textless moral\textgreater"\\
\}\} \\
\end{tcolorbox}

%%% MORAL KEYWORD
% Moral Keyword Prompt, again 8 variations, 2 phrasings x before or after the article text x with or without role statement
\noindent \textbf{Moral Keyword Role Statement:}
\begin{tcolorbox}[colback=gray!10, colframe=gray!10, boxrule=0pt]
\small
Today, you are an expert story interpreter. I will give you a sentence summary of a news article and ask you a question about it.
\end{tcolorbox}

\noindent \textbf{Moral Keyword Prompt, Phrasing 1:}
\begin{tcolorbox}[colback=gray!10, colframe=gray!10, boxrule=0pt]
\small
Label the sentence with a fitting moral keyword or keywords. Use only one word if appropriate, or up to three distinct words if multiple moral ideas are clearly present. Avoid using words like 'technology' or 'AI'. \\
Please respond only with a valid JSON object in the following format: \\
\}\} \\
"MORAL\_KEYWORD\_1": "\textless keyword\textgreater", \\
"MORAL\_KEYWORD\_2": "\textless keyword\textgreater", \\
...\\
\}\} \\
\end{tcolorbox}

\noindent \textbf{Moral Keyword Prompt, Phrasing 2:}
\begin{tcolorbox}[colback=gray!10, colframe=gray!10, boxrule=0pt]
\small
Can you reduce this to between one and three moral keywords? Use only one word if appropriate, or up to three distinct words if multiple moral ideas are clearly present. Avoid using words like 'technology' or 'AI'. \\
Please respond only with a valid JSON object in the following format: \\
\}\} \\
"MORAL\_KEYWORD\_1": "\textless keyword\textgreater", \\
"MORAL\_KEYWORD\_2": "\textless keyword\textgreater", \\
...\\
\}\} \\
\end{tcolorbox}

%% if required, the content of .bbl file can be included here once bbl is generated
%%\input sn-article.bbl

\end{document}